\pdfoutput=1

\documentclass[11pt]{article}

\usepackage[]{EACL2023}

\usepackage[T2A]{fontenc} 
\usepackage{hyphenat}
\usepackage[english, russian]{babel}

\usepackage{times}
\usepackage{latexsym}

\usepackage{breqn}
\usepackage{amsmath}
\usepackage{url}

\usepackage[T1]{fontenc}

\usepackage[utf8]{inputenc}

\usepackage{comment}
\usepackage{microtype}

\usepackage{soul}
\usepackage{epstopdf}
\usepackage{booktabs}
\usepackage{graphicx}

\usepackage{inconsolata}

\title{Investigating Massive Multilingual Pre-Trained Machine Translation Models for Clinical Domain via Transfer Learning}

\author{
          Lifeng Han$ ^1$, Gleb Erofeev$^{ 2}$,    Irina Sorokina$^{ 2}$,  Serge Gladkoff$ ^2$, 
\and \textbf{Goran Nenadic}$^{1}$ \\
         $^1$ The University of Manchester, UK \\ 
         $^2$ Logrus Global,  Translation \& Localization 
         \\ {\tt lifeng.han, g.nenadic@manchester.ac.uk} 
         \\
         {\tt 
         gleberof, irina.sorokina, serge.gladkoff@logrusglobal.com}         }

\begin{document}
\maketitle
\begin{abstract}
Massively multilingual pre-trained language models (MMPLMs) are developed in recent years demonstrating superpowers and the pre-knowledge they acquire for downstream tasks.
This work investigates whether MMPLMs can be applied to clinical domain machine translation (MT) towards entirely unseen languages via transfer learning.
We carry out an experimental investigation using Meta-AI's MMPLMs ``wmt21-dense-24-wide-en-X and X-en (WMT21fb)'' which were pre-trained on 7 language pairs and 14 translation directions including English to Czech, German, Hausa, Icelandic, Japanese, Russian, and Chinese, and the opposite direction.
We fine-tune these MMPLMs towards English-\textit{Spanish} language pair which \textit{did not exist at all} in their original pre-trained corpora both implicitly and explicitly.
We prepare carefully aligned \textit{clinical} domain data for this fine-tuning, which is different from their original mixed domain knowledge.
Our experimental result shows that the fine-tuning is very successful using just 250k well-aligned in-domain EN-ES segments for three sub-task translation testings: clinical cases, clinical terms, and ontology concepts. It achieves very close evaluation scores to another MMPLM NLLB from Meta-AI, which included Spanish as a high-resource setting in the pre-training.
To the best of our knowledge, this is the first work on using MMPLMs towards \textit{clinical domain transfer-learning NMT} successfully for totally unseen languages during pre-training. 
\end{abstract}

\section{Introduction}
\label{section_intro}
Multilingual neural machine translation (MNMT) has its root from the beginning of NMT era \cite{dong-etal-2015-multiTask_NMT,firat-etal-2016-multiWay_MNMT}
but only made its first milestone when Google's end-to-end MNMT arrived \cite{johnson-etal-2017-googles} where the artificial token was introduced for the first time for translation task at the beginning of the input source sentence to indicate the specified target language, e.g. ``2en'' as translating into English. 
This model used a shared word-piece vocabulary and enabled multilingual NMT through a single encoder-decoder model training. Google's MNMT also demonstrated the possibility of ``zero-shot'' translation as long as the languages to be translated from or to have been seen during the training stage, even though not explicitly. 
However, as the authors mentioned, Google's MNMT \textit{only} allows translating between languages that have been seen individually as ``source and target languages during some point, not for entirely new ones'' in their many-to-many model, which was tested using the WMT14 and WMT15 data \cite{johnson-etal-2017-googles}. 
This set an obstacle to translating freshly new languages that do not exist in their pre-training stage. 
Then using the later developed NMT structure Transformer and BERT \cite{devlin-etal-2019-bert,google2017attention}, Facebook AI extended the coverage of multilingual translation into 50, 100, and 200+ languages via mBERT-50 \cite{DBLP:journals/corr/abs-2008-00401_mBERT-50}, M2M-100 \cite{10.5555/3546258.3546365_m2m_100}, and NLLB \cite{https://doi.org/10.48550/arxiv.2207.04672_NLLB} models. 
However, these models never address the issue of translating entirely new languages that do not exist in their pre-training stage, which sets an obstacle for MT applications in serving an even broader community.

In this work, we move one step forward towards domain-specific \textit{transfer-learning} \cite{zoph-etal-2016-transfer} for NMT via fine-tuning an entirely new language pair that does not exist in the deployed multilingual pre-trained language models (MPLMs). 
The MPLMs we used are from Facebook AI (Meta-AI)' s submission to the WMT21 news translation task, i.e. ``wmt21-dense-24-wide-en-X'' and ``wmt21-dense-24-wide-X-en'' which were pre-trained for 7 languages Hausa (ha), Icelandic (is), Japanese (ja), Czech (cs), Russian (ru), Chinese (zh), German (de) to English (en), and backward \cite{tran2021facebook}. 
We use a well-prepared 250k pairs of English-Spanish (en-es) clinical domain corpus and demonstrate that not only it is possible to achieve successful transfer-learning on this explicit new language pair, i.e. the Spanish language is totally unseen among the languages in the MPLM, but also the domain knowledge transfer from general and mixed domain to the clinical domain is very successful. 
In comparison to the massively MPLM (MMPLM) NLLB which covers Spanish as a high-resource language at its pre-training stage, our transfer-learning model achieves very close evaluation scores in most sub-tasks (clinical cases and clinical terms translation) and even wins NLLB in ontology concept translation task by the metric COMET \cite{rei-etal-2020-comet} using ClinSpEn2022 testing data at WMT22. This is a follow-up work reporting further findings based on our previous shared task participation \cite{han-etal-2022-examining}.

\begin{figure*}[!t]
\centering
\includegraphics*[width=0.85\textwidth]{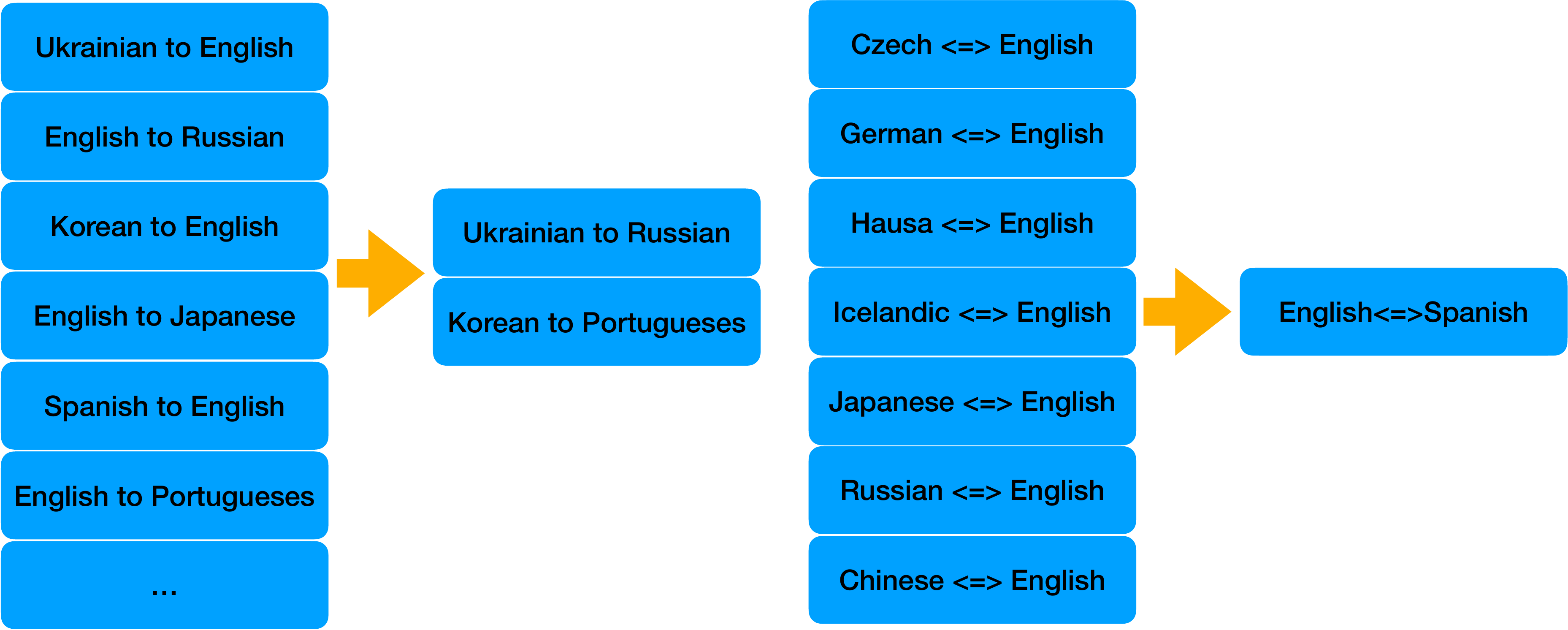}
\caption{(Figure:) Difference of Google's Multi-lingual NMT Bridge Models (left) and Our Transfer-Learning Model (right).}
\label{fig:diff_googleMNMT_ours}
\end{figure*}

\section{Related Work}
\label{section_related_work}

Regarding the early usage of special tokens in NMT, \newcite{sennrich-etal-2016-controlling_politeNMT} designed the token T from Latin \textit{Tu} and V from Latin \textit{Vos} for familiar and polite indicators attached to the source sentences towards English-to-German NMT. \newcite{yamagishi-etal-2016-controlling_voiceNMT} designed tokens <all-active>, <all-passive>, <reference> and <predict> to control of voice of Japanese-to-English NMT; either they are active, passive, reference aware or prediction guided.
Subsequently, Google's MNMT system designed target language indicators, e.g. <2en> and <2jp> controlling the translation towards English and Japanese respectively \cite{johnson-etal-2017-googles}. 
Google's MNMT also designed mixed target language translation control, e.g.  (1-$\alpha$)<2ko> + $\alpha$<2jp> tells a mixed language translation into Korean and Japanese with a weighting mechanism.
We take one step further to use an existing language controller token from a MPLM as a \textit{pseudo code} to fine-tune an \textit{external language} translation model, which was entirely not seen during the pre-training stage.

Regarding transfer-learning applications 
for downstream NLP tasks other than MT, \newcite{muller-etal-2021-unseen_MLM} applied transfer learning from MPLMs towards unseen languages of different typologies on dependency parsing (DEP), named entity recognition (NER), and part-of-speech (POS) tagging. \newcite{ahuja-etal-2022-multiTask_zeroshotMM} carried out zero-shot transfer learning for natural language inference (NLI) tasks such as question answering.

In this paper, we ask this research question (RQ):
\textit{Can Massive Multilingual Pre-Trained Language Models Create a Knowledge Space Transferring to Entirely New Language (Pairs) and New (clinical) Domains for Machine Translation Task via Fine-Tuning?}

\section{Model Settings}
\label{section_model_setting}
To investigate into our RQ, we take Meta-AI's MNMT submission to WMT21 shared task on news translation, i.e. the MMPLM ``wmt21-dense-24-wide-en-X'' and ``wmt21-dense-24-wide-X-en'' as our test-base, and we name them as WMT21fb models \cite{tran2021facebook}\footnote{\url{https://github.com/facebookresearch/fairseq/tree/main/examples/wmt21}}. 
They are conditional generation models from the same structure of massive M2M-100 \cite{10.5555/3546258.3546365_m2m_100} having a total number of 4.7 billion parameters which demand high computational cost for fine-tuning. 
WMT21fb models were trained on mixed domain  data using ``all available resources'' they had, for instances, from historical WMT challenges, large-scale data mining, and their in-domain back-translation. Then these models were fine-tuned in news domain for 7 languages including Hausa, Icelandic, Japanese, Czech, Russian, Chinese, German from and to English. 


The challenging language we choose is Spanish, which did not appear in the training stage of WMT21fb models.
The fine-tuning corpus we use is extracted from MeSpEn \cite{villegas2018mespen} clinical domain data, of which we managed to extract 250k pairs of English-Spanish segments  after data cleaning. 
They are from 
IBECS-descriptions,
IBECS-titles,
MedlinePlus-health\_topics-titles,
MedlinePlus-health\_topics-descriptions,
Pubmed-descriptions,
Scielo-descriptions, and 
Scielo-titles. 

To implement the fine-tuning, we use the <2en> token for translating from Spanish to English, and <2ru> (originally to Russian) \textit{pseudo token} for translating towards English-to-Spanish (en2es) \footnote{using <2es> token will result into errors since Spanish was actually not used in the WMT21fb PLMs}.
The difference between our transfer-learning NMT model and Google's MNMT can be shown in Figure \ref{fig:diff_googleMNMT_ours}, right \textit{vs} left. 
In Google's MNMT model, it can only translate ``new language pairs'' that are not explicitly seen but implicitly seen, e.g. bridging language pairs (Ukrainian-to-English and English-to-Russian $\Rightarrow$ Ukrainian-to-Russian), or language pairs that have been seen as source (Korean) and target (Portuguese) somewhere. 
In our transfer-learned NMT, Spanish was not among the trained languages at all.

In comparison, we deploy another MMPLM from Meta-AI, i.e. the ``No-Language-Left-Behind (NLLB)'' which was trained on 204 languages including Spanish as one of their high-resource ones \cite{https://doi.org/10.48550/arxiv.2207.04672_NLLB}. NLLB full model is a massive size Transformer having 55 billion parameters and we use its distilled version NLLB-200-distilled \footnote{\url{https://huggingface.co/facebook/nllb-200-distilled-1.3B}}, which still has 1.3 billion parameters.
Fine-tuning is carried out on NLLB using the same 250K ES-EN corpus.

\begin{table*}[!th]
\begin{center}
\centering
\begin{tabular}{ccccccc}
\toprule
\multicolumn{1}{c}{} 
     & \multicolumn{5}{c}{Task-I: Clinical Cases (CC) EN$\rightarrow$ES}    \\ \hline 
\multicolumn{1}{c}{MT fine-tuning} 
     & in.es? & \multicolumn{1}{c}{S\textsc{acre}BLEU}     
                & METEOR & COMET & BLEU-HF & ROUGE-L-F1 \\
\midrule
Clnical-NLLB & Yes & 37.74 & 0.6273& 0.4081& 0.3601& 0.6193 \\
Clinical-WMT21fb & \textbf{No} &34.30 &0.5868 &0.3448 &0.3266 & 0.5927\\
\hline\hline
\multicolumn{1}{c}{} 
     & \multicolumn{5}{c}{Task-II: Clinical Terms (CT)  EN$\leftarrow$ES }    \\ \hline 
\multicolumn{1}{c}{MT fine-tuning} 
     & in.es? & \multicolumn{1}{c}{S\textsc{acre}BLEU}     
                & METEOR & COMET & BLEU-HF & ROUGE-L-F1 \\
\midrule
Clinical-NLLB & Yes& {28.57} & 0.5873 & {1.0290} & {0.2844} & 0.6710 \\
Clinical-WMT21fb & \textbf{No} &24.39 &0.5840 &0.8584 & 0.2431 & 0.6699\\
 \hline\hline
\multicolumn{1}{c}{} 
     & \multicolumn{5}{c}{Task-III: Ontology Concept (OC)  EN$\rightarrow$ES  }    \\ \hline
\multicolumn{1}{c}{MT fine-tuning} 
     & in.es? & \multicolumn{1}{c}{S\textsc{acre}BLEU}     
                & METEOR & COMET & BLEU-HF & ROUGE-L-F1 \\
\midrule
Clinical-NLLB & Yes& {41.63}  & 0.6072& 0.9180& {0.3932} & 0.7477 \\
Clinical-WMT21fb& \textbf{No} &40.71 &0.5686  & \textit{0.9908} & 0.3859 & 0.7199 \\
\bottomrule
\end{tabular}
\caption{(Table:) Evaluation Scores using Five Official Metrics from ClinSpEn2022 Benchmark on Two Models. The column ``in.es'' means if the original pre-trained model included the Spanish language before fine-tuning/transfer-learning.}
\label{tab:clinSpEn_eval_score_t123}
\end{center}
\end{table*}

\section{Model Evaluations}
\label{section_wmt22}

\subsection{Testing Corpus from Clinical Domain}
We used the official testing corpus from ClinSpEn2022 shared task affiliated to Biomedical-MT at WMT22. ClinSpEn2022 aims at developing clinical domain machine translation on Spanish-English language pair\footnote{\url{https://temu.bsc.es/clinspen/}}, which is hosted in CodaLab \cite{codalab_competitions} \footnote{\url{https://codalab.lisn.upsaclay.fr/competitions/6696}}. 

There are three sub-tasks: 1) Clinical Cases (CC): on 202 COVID-19 clinical case reports; 2) Clinical Terms (CT): using more than 19K parallel terms extracted from biomedical literature and electric health records (EHRs); 3) Ontology Concepts (OC): using more than 2K parallel concepts from biomedical ontology. 
The translation direction on these three sub-tasks are EN$\rightarrow$ES, EN$\leftarrow$ES, and EN$\rightarrow$ES respectively.

\subsection{Evaluation Metrics}
The official evaluation metrics used by ClinSpEn2022 shared task are METEOR \cite{BanerjeeLavie2005}, S\textsc{acre}BLEU \cite{post-2018-call4clarity}, COMET \cite{rei-etal-2020-comet}, BLEU-HF (HuggingFace) \cite{papineni-etal-2002-bleu}, and ROUGE-L-F1 \cite{lin-2004-rouge}. Among these, METEOR is a metric using both precision and recall not only on word surface level but also introducing paraphrasing features. COMET was proposed recently by taking advantage of cross-lingual PLMs using knowledge from both source and target languages. ROUGE was originally designed for text summarisation evaluation using n-gram co-occurrences, while ROUGE-L
added the Longest Common Subsequence (LCS) feature from translation study.

The reporting of BLEU metric scores has certain uncertainty, which is caused by some parameter settings when using BLEU metric including number of references, length penalty computation on multi-references, maximum n-gram, and smoothing applied to 0-count n-grams. 
To address these issues, S\textsc{acre}BLEU added some constrains while using BLEU metric. These include the applying of its own metric-internal pre-processing for detokenised system outputs, the avoiding of user handling reference set via automatically downloading from WMT, and the export of a summary on settings used.

\subsection{Evaluation Scores}

We present the MT evaluation scores using five official metrics from ClinSpEn2022 shared task on the three sub-tasks in Table \ref{tab:clinSpEn_eval_score_t123}, for translating clinical cases, clinical terms, and clinical concepts. 
The two fine-tuned models are clinic-NLLB which is achieved by domain fine-tuning and clinic-WMT21fb which is a domain fine-tuning plus transfer-learning model to a new language space.

On Task 1 and 2, Clinical-WMT21fb has very comparable evaluation scores to clinical-NLLB, even though it only used 250k pairs es-en sentences for fine tuning without seeing any en-es or Spanish language at all during pre-training. In contrast, clinical-NLLB used a large amount of Spanish data for its pre-training phase.
On Task 3, the evaluation scores of these two models are even closer on BLEU and S\textsc{acre}BLEU, especially the clinical-WMT21fb wining COMET metric over clinical-NLLB (0.9908 \textit{vs} 0.9180).



This experimental result shows that with a carefully prepared certain amount of fine-tuning data, e.g. 250k pair of sentences, the MMPLMs are capable to create a semantic knowledge space transferring to an entirely new (external) language pair for NMT task in a new domain, i.e. clinical domain. This answers our RQ set up in the beginning of this investigation.


\subsection{Human Evaluation}
We looked into three sub-task translation outputs from the model clinical-WMT21fb. It shows that for the  EN$\leftarrow$ES translation task, i.e. the sub-task 2 clinical term translation, the output file is totally file with only English tokens. 
On the other two sub-tasks, i.e. the clinical cases and ontology concept translation, which have the translation direction  EN$\rightarrow$ES, there are some Russian tokens in the output, not only Spanish tokens. 
However, the Russian tokens in the Spanish sentences are not nonsense, instead proper translations of entities and words. 
The entire test set of these two sub-tasks is very large around 300K sentences/segments, and there are only 12K lines of them (4\%) have Russian tokens. 
So we have fine-tuned the model in EN-RU direction on EN-SP data, and it translates well into Spanish! But if there isn’t a suitable Spanish token in the generation model,  it takes a Russian token.

We also looked into the translation outputs from clinic-NLLB model for error analysis using two native Spanish speakers, one of them having a PhD degree in biomedical NLP field and the other having a Master degree in translation studies.
The error analysis shows that some of the translation errors come from very literal translation, and others come from gender related mistakes. This suggests that the massively pre-trained MLM is still not there to capture the differences of linguistic features among pre-trained languages.

\section{Discussion}

\subsection{On Automatic Metrics}
We had more thoughts on the automatic evaluation settings and outputs, especially on the COMET metric in comparison to others.

Firstly, the closeness of most automatic metric scores does not necessarily mean that the translation outputs are very good.
Most metrics only measure the linguistic proximity of outputs to the ``gold standard of reference''.

Secondly, COMET is a reference-less metric taking advantage of cross-lingual PLMs using knowledge from both source and target languages. This has pros and cons: a) it might be able to capture the semantic relatedness without seeing the same language tokens, even in the same sequence/sentence; b) also due to this, it is not able to distinguish foreign language tokens in the translation output, which normally shall receive a penalty in evaluation scores. This also inspires another research topic, i.e. shall we really punish the foreign or mixed-language tokens in the translation output in all evaluation conditions, or it shall depend on the situation of the output applications? This has an echo to Google's zero-shot MNMT model \cite{johnson-etal-2017-googles} when the mixed language tokens are used for translation model, e.g. (1-$\alpha$)<2KO>+$\alpha$<2JP> resulting in mixed tokens of Korean and Japanese in the output translation but they are \textit{semantically correct tokens}.

In a situation when users want only the Spanish translation output, 4\% of Russian tokens in the Spanish translation should surely receive a penalty in the quality evaluation setting. The COMET metric will fail this mission, and professional human evaluation is always much needed for trustworthiness.
However, in a situation to measure the models' cross-lingual capability on semantic preservation for direct output, or as input into other ML models, is it better to generate \textit{NULL} or meaningless tokens or random translations in the target language, or to choose semantically correct foreign tokens when the model does not know how to predict the exact correct target tokens? This inspires us to think again about the evaluation setting on different tasks.

\subsection{MT System Output Examples}
\begin{figure*}[!th]
\centering
\includegraphics*[width=0.99\textwidth]{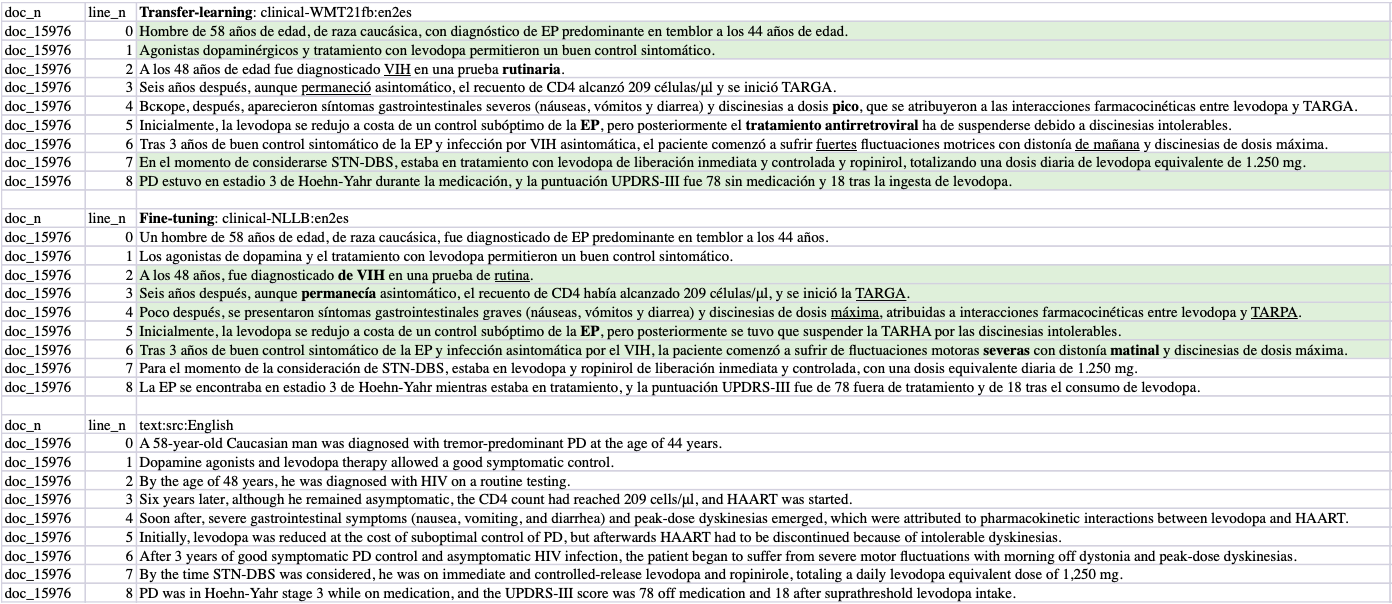}
\caption{(Figure:) Task-1 Cases/Sentences EN-ES Translation Examples: clinic-WMT21fb \textit{vs} clinic-NLLB}
\label{fig:cases_task_example}
\end{figure*}

\begin{figure*}[!th]
\centering
\includegraphics*[width=0.99\textwidth]{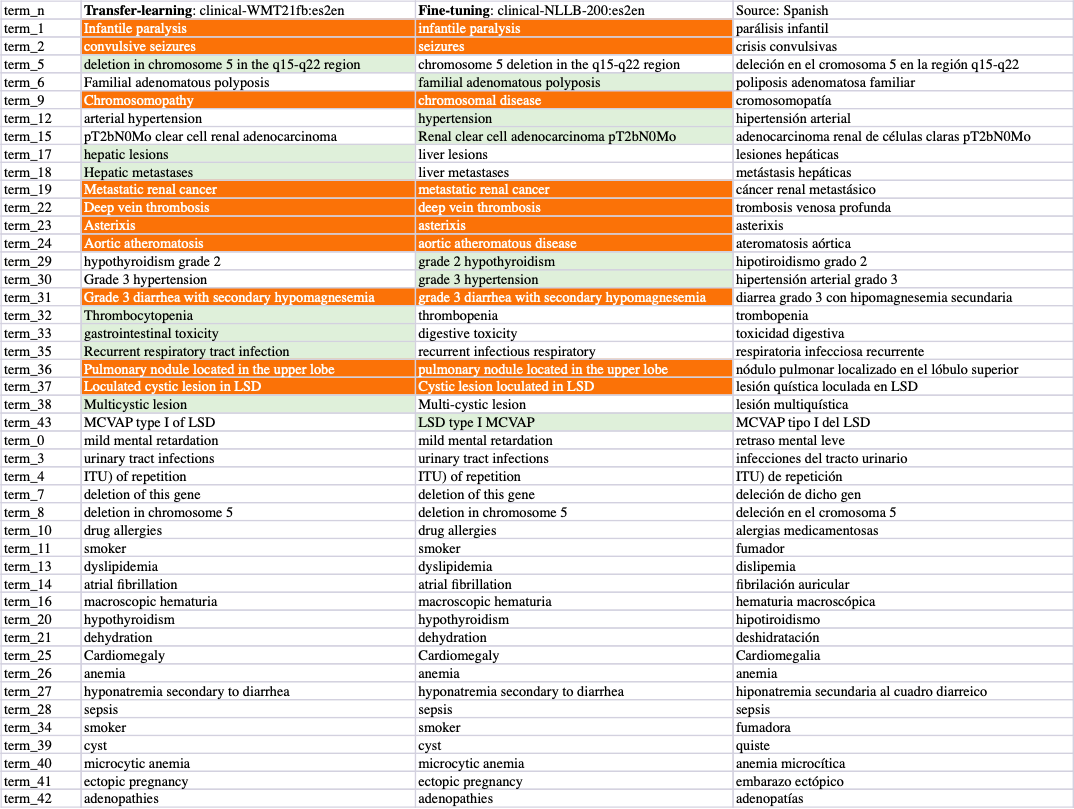}
\caption{(Figure:) Task-2 Clinical Term ES-EN Translation Examples: clinic-WMT21fb \textit{vs} clinic-NLLB}
\label{fig:term_task_example}
\end{figure*}

\begin{figure*}[!th]
\centering
\includegraphics*[width=0.99\textwidth]{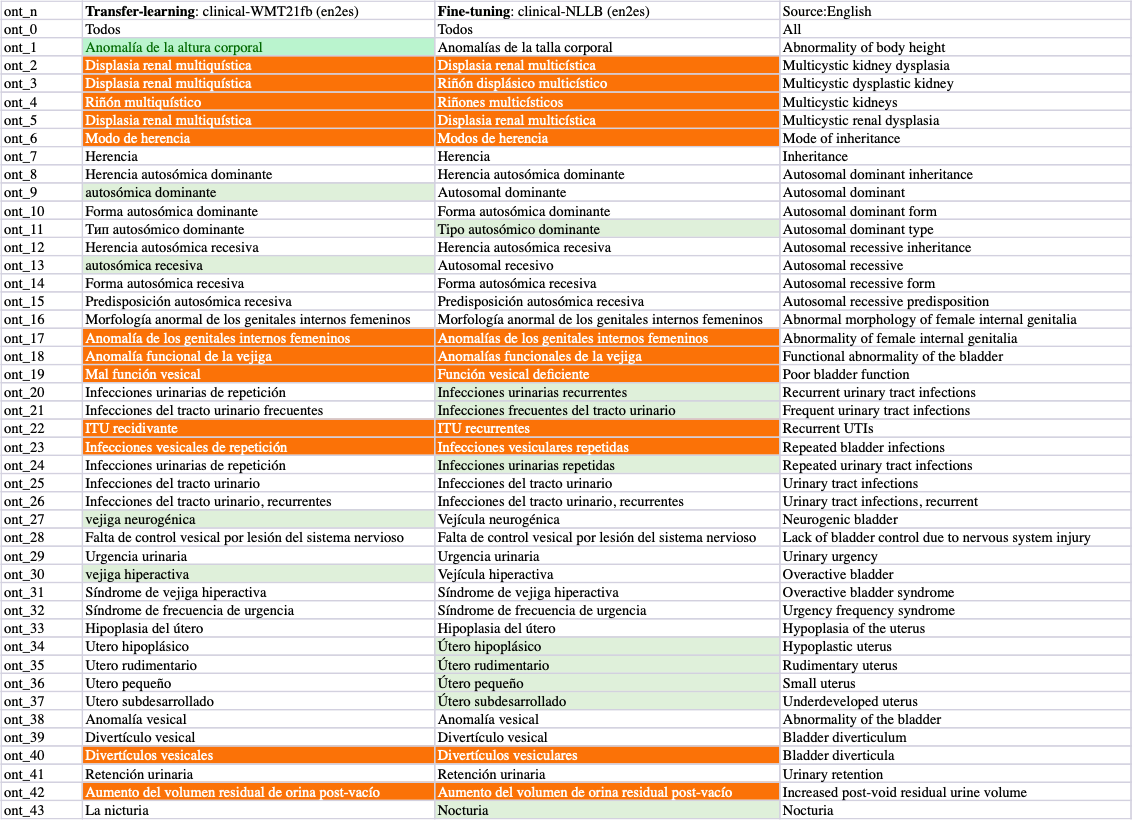}
\caption{(Figure:) Task-3 Concept EN-ES Translation Examples: clinic-WMT21fb \textit{vs} clinic-NLLB}
\label{fig:concept_task_example}
\end{figure*}

We present the MT system output examples from both  clinical-WMT21fb and  clinical-NLLB-200 for three tasks in Figure \ref{fig:cases_task_example}, \ref{fig:term_task_example}, and \ref{fig:concept_task_example}. 
In these figures, the green colour is for the ``preferred translations'' while the orange colour is for ``both sounds good''. 
The annotations were firstly marked by one of the two human evaluators we have, and then verified by the second native Spanish speaker.

From these sampled MT outputs, the model clinical-WMT21fb sometimes outperforms clinical-NLLB-200, and vice versa. For instance, in the concept translation (Figure \ref{fig:concept_task_example}), the English concept ``Abnormality of body height'' (ont\_1) is better translated by transfer-learned model into ``Anomalía de la altura corporal'' than ``Anomalías de la talla corporal'' by clinical-NLLB, since ``altura'' means ``height'' while ``talla'' actually means ``size'' which is not accurate. We will carry out a systematic human evaluation in a larger sample size.

Regarding rare Russian tokens from the language-transferred model, in Task-1, ``Вскоре'' from clinical-WMT21fb in line\_n 4 means ``soon'', even though it is a Russian token, i.e. non-Spanish token. 
In Task-3, ``Тип’’ in ``Тип autosómico dominante'' means ``type of'' from ont\_11 which is a meaningful Russian token. 

\section{Conclusion and Future Work}
\label{section_discussion}
We investigated if \textit{real} transfer-learning NMT is possible using massive multilingual pre-trained LMs (MMPLMs) to translate \textit{external} languages that are unseen \textit{at all} in the training phase. 
We used Meta-AI's mixed domain multilingual PLMs (WMT21fb) as our test base, 250K well-prepared EN-ES clinical data as fine-tuning corpus, and <2ru> as pseudo-code for new language (out-of-en) fine-tuning. 
We tested the fine-tuned model on ClinSpEn2022 clinical domain shared task data, and the results show that this fine-tuning is successful, which achieves very comparable scores to Meta-AI's MMPLM NLLB model, which had Spanish in the training phase as a high-resource setting.
We think this demonstrates that the Hyper-Transformer model from WMT21fb does build a language-independent ``semantic space'' that allows one to understand a different language and correctly construct a totally different language model when fine-tuned on the language which is absent and different from the languages it was trained upon. 
There are many future works that can be carried out based on the findings from this work. 
Firstly, we plan to carry out an extensive human-expert-based evaluation, e.g. using HOPE metric \cite{gladkoff-han-2022-hope}, looking into the differences between the outputs of these two MMPLMs, such as on translating multi-word expressions in the clinical domain \cite{bhatia2023proceedings,han2022investigation}. We also designed corresponding measurements on the evaluation of uncertainty and inter-rater reliability (IRR) levels \cite{gladkoff-etal-2022-measuring,gladkoff2023student}. 
Secondly, we think it is valuable to integrate more high-performance automatic metrics into the comparison such as hLEPOR \cite{han2021cushlepor}.
Finally, we will try more external languages from different typologies in future work.

\subsection*{Limitations}
1) \textbf{\textit{On PLM Capability for Transferring to New Language}},
 in this work, we used Meta-AI's WMT21 multilingual pre-trained language models as our test-base for the knowledge transfer into an external language fine-tuning and translation.
This new-language ability is much dependent on the MPLMs we used, such as WMT21fb \cite{tran2021facebook} as a huge size model, a conditional generation from Meta-AI's massive M2M-100 model \cite{10.5555/3546258.3546365_m2m_100}. 
If we try to fine-tune a \textit{bilingual model} on an external language that the PLM did not see, it will not be that good because for smaller-sized models such fine-tuning would be too much of a change, and the model will lose generalisation which leads to problems.
For huge multilingual PLM models, the 250K of fine-tuning data is a small set of numbers, and that's why the model does not lose generalisation and captures new data well without losing linguistic knowledge of other languages that it was trained on.

\noindent 2) \textbf{On the Impact of Language Families}, the MMPLM WMT21fb we deployed has both alphabetic languages and CJK (\textit{Chinese}, \textit{Japanese}, Koran) character languages, as well as Slavic language (Russian). This might make it easier to transfer to a new language, e.g. alphabetic language. However, in situations when the MPLMs did not include any of the language scripts that belong to the language family of the target one, it can be much harder for it to transfer to the new target language. This needs further investigation. One possible extension of this work is using the dynamic vocabulary method proposed by \newcite{lakew-etal-2018-transfer}.

\section*{Acknowledgements}
The authors thank the ClinSpEn2022 shared task organisers for preparing the data set and evaluation platforms, thank Darryl Estrada for communicating with us during the competition.
We thank Dr Alfredo Madrid Garcia and Ms Cristina Sánchezfor carrying out the human evaluation for this work.
We thank the open research projects
Meta-AI's wmt21.dense-24-wide.En-X (2021) and NLLB (2022) we used.
This work has been partially supported by grant EP/V047949/1 "Integrating hospital outpatient letters into the healthcare data space" (funder: UKRI/EPSRC).

\begin{figure*}[t]
\centering
\includegraphics*[width=0.85\textwidth]{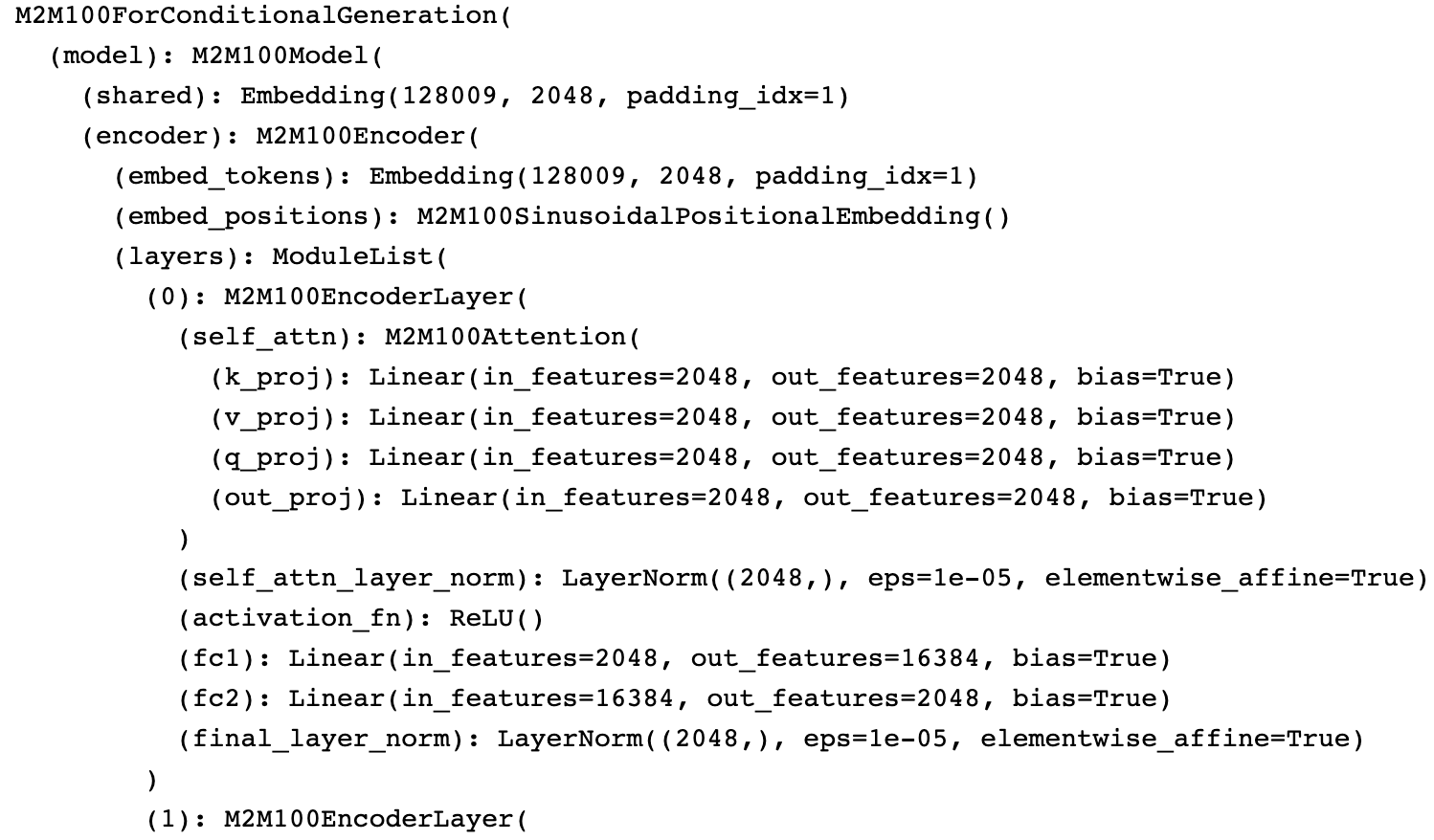}
\caption{M2M-100 Model Structure For Conditional Generation Encoder: Samples and Parameters}
\label{fig:M2M_100_encoder}
\end{figure*}

\bibliography{anthology,custom}
\bibliographystyle{acl_natbib}

\appendix

\section{Appendices}
\label{sec:appendix}

\subsection{Model Parameters in Detail}
\noindent Some fine-tuning parameters for NLLB-200-distilled \cite{https://doi.org/10.48550/arxiv.2207.04672_NLLB} are listed below:
\begin{itemize}
    \item batch size = 24
    \item gradient accumulation steps = 8
    \item weight decay = 0.01
    \item learning rate = 2e-5
    \item number of training epochs = 1
    \item encoder-decoder layers = 24+24
    \item Activation function (encoder/decoder) = ReLU

\end{itemize}

The Parameters for fine-tuning WMT21fb model are the same as for the NLLB-200, except for the batch size which is set as 2, which is because the model is too large and we got an OOM error if the batch size is set above 2.
\noindent More details on M2M-100 for Conditional Generation structure \cite{10.5555/3546258.3546365_m2m_100} we used can be find in Figure \ref{fig:M2M_100_encoder}.

\end{document}